\documentclass[journal]{IEEEtran}

\ifCLASSINFOpdf
  \usepackage[pdftex]{graphicx}
  \graphicspath{{./figs/}}
  \DeclareGraphicsExtensions{.pdf,.jpeg,.png}
\else
  \usepackage[dvips]{graphicx}
  \graphicspath{{./figs/}}
  \DeclareGraphicsExtensions{.eps}
\fi

\usepackage[cmex10]{amsmath}     % loads amsmath with Type1 fonts
\usepackage{amssymb}
\usepackage{amsfonts}
\usepackage{mathtools}
\usepackage{bm}
\usepackage{pifont}
\usepackage{array}
\usepackage{booktabs}
\usepackage{stfloats}
\usepackage{multirow}
\usepackage{cite}
\usepackage[table]{xcolor}
\usepackage{booktabs}
\usepackage{algorithm}
\usepackage{algpseudocode}

\usepackage[caption=false,font=footnotesize]{subfig}

\usepackage{url}
\usepackage[colorlinks,linkcolor=blue,citecolor=blue,urlcolor=blue]{hyperref}

\begin{document}

% \title{Hierarchical U-Net Framework for Enhanced Urban Flood Mapping in Arid Environments Using PlanetScope Imagery}

\title{FM-LC: A Hierarchical Framework for Urban Flood Mapping by Land Cover Identification Models}

\author{Xin~Hong*,
        Longchao~Da*\footnote,~\IEEEmembership{Member,~IEEE, Hua~Wei, Member,~IEEE}
\thanks{This work has been submitted to the IEEE for possible publication. Copyright may be transferred without notice, after which this version may no longer be accessible.}
\thanks{$^*$Authors contributed equally to this work. Xin Hong is with the Department of Environmental Sciences, College of Natural and Health Sciences, Zayed University, the United Arab Emirates and Longchao Da and Hua Wei is with the Computer Science department of Arizona State University, Tempe, USA. (e-mail: xin.hong@zu.ac.ae; longchao@asu.edu, hua.wei@asu.edu).}
}

\markboth{IEEE Transactions on Geoscience and Remote Sensing,~Vol.~XX, No.~XX, Month~2025}%
{Hong and Da: Hierarchical U-Net for Flood Mapping}

\maketitle

\begin{abstract}
% This study presents a hierarchical deep learning framework for enhanced urban flood mapping in arid environments using high-resolution PlanetScope imagery. Building upon a previously trained U-net model, we propose a three-stage architecture that includes a lightweight binary flood detector, a multi-class land use/land cover (LULC) segmentation model, and a Bayesian smoothing module for spatial refinement. This hierarchical approach improves segmentation accuracy by isolating flood-prone zones and refining boundaries between LULC classes with similar spectral signatures. The method was applied to the April 2024 extreme rainfall event in Dubai, UAE, achieving significant improvements in classification metrics compared to the baseline U-net model. Results show a XX\% increase in F1-scores across all classes with notable gains in flood and vegetation detection. The proposed framework demonstrates strong potential for operational flood monitoring in urban arid regions and advances the use of deep learning for fine-scale, multi-temporal geospatial analysis.

Urban flooding in arid regions poses severe risks to infrastructure and communities. Accurate, fine‐scale mapping of flood extents and recovery trajectories is therefore essential for improving emergency response and resilience planning. However, arid environments often exhibit limited spectral contrast between water and adjacent surfaces, rapid hydrological dynamics, and highly heterogeneous urban land covers, which challenge traditional flood‐mapping approaches. High‐resolution, daily PlanetScope imagery provides the temporal and spatial detail needed. In this work, we introduce FM-LC, a hierarchical framework for \textbf{\underline{F}}lood \textbf{\underline{M}}apping by \textbf{\underline{L}}and \textbf{\underline{C}}over identification, for this challenging task. Through a three-stage process, it first uses an initial multi-class U-Net to segment imagery into water, vegetation, built area, and bare ground classes. We identify that this method has confusion between spectrally similar categories (e.g., water vs. vegetation). Second, by early checking, the class with the major misclassified area is flagged, and a lightweight binary `expert' segmentation model is trained to distinguish the flagged class from the rest. Third, a Bayesian smoothing step refines boundaries and removes spurious noise by leveraging nearby pixel information. We validate the framework on the April 2024 Dubai storm event, using pre- and post-rainfall PlanetScope composites. Experimental results demonstrate average F1-score improvements of up to 29\% across all land-cover classes and notably sharper flood delineations, significantly outperforming conventional single-stage U-Net baselines.

\end{abstract}
 

\IEEEpeerreviewmaketitle

\section{\textbf{Introduction}}
% \IEEEPARstart{T}{his} demo file is intended to serve as a starter file for IEEE journal papers produced under \LaTeX. Replace with your own introduction.
\IEEEPARstart{U}{rban} flooding poses significant challenges to public safety and the functioning of cities, especially in rapidly developing metropolitan regions located in arid environments where drainage infrastructure typically lacks the capacity to handle extreme rainfall events \cite{almheiri2023review}. The heavy rainfall in Dubai, United Arab Emirates (UAE), in April 2024 marked the most severe rainstorm in 75 years. This extreme downpour exposed key vulnerabilities in the Persian Gulf region, such as inadequate drainage infrastructure and limited urban preparedness, which led to widespread flooding and significant infrastructure and socioeconomic disruptions \cite{alawlaqi2025we, cornish2024dubai, oxfordanalytica2024flash}. These events underscore the overarching need for robust, high-resolution flood mapping coupled with dynamic post-event monitoring to inform proactive resilience strategies and optimize emergency response operations.

To meet these needs, PlanetScope imagery that offers high spatial (3 m) and daily temporal resolutions emerges as an ideal data source for detailed urban flood mapping and post‐event recovery assessments. Several recent studies have shown that PlanetScope outperforms coarser‐resolution sensors such as Sentinel-2 and Landsat in accurately delineating inundated areas and quantifying flood impacts \cite{ibrahim2024evaluating, alawathugoda2024impact}. However, most existing workflows for PlanetScope rely on conventional machine‐learning classifiers, which demand extensive manual feature engineering and careful parameter tuning. Such approaches often struggle to generalize across diverse environments and fail to exploit the full potential of high‐frequency, high‐resolution data \cite{alawathugoda2024impact, lefulebe2022fine}.

% Deep learning techniques, particularly convolutional neural networks (CNNs) such as U-Net, have shown substantial improvements over traditional classifiers by automatically extracting intricate spatial patterns from remote sensing images \cite{digra2022land, valman2024ai, qayyum2020glacial}. Despite their effectiveness, deep learning applications on PlanetScope imagery have mainly focused on single-class scenarios, such as water bodies, without addressing comprehensive multi-class land use and land cover (LULC) classifications critical for holistic urban flood assessments.

Deep learning models, particularly the encoder–decoder architectures such as the U-Nets, excel at learning multi‐scale spatial features~\cite{da2025deepshade} for remote‐sensing segmentation \cite{digra2022land, valman2024ai, qayyum2020glacial}. However, when applied to high-resolution PlanetScope imagery, standard U-Net implementations often confuse spectrally similar classes (e.g., inundated surfaces vs.\ moist vegetation), resulting in misclassified flood extents. Addressing these spectral ambiguities requires a specialized, multi-stage framework that can both perform broad multi-class segmentation and then refine the most error-prone regions.

\begin{figure}[t!]
    \centering
    \includegraphics[width=0.82\linewidth]{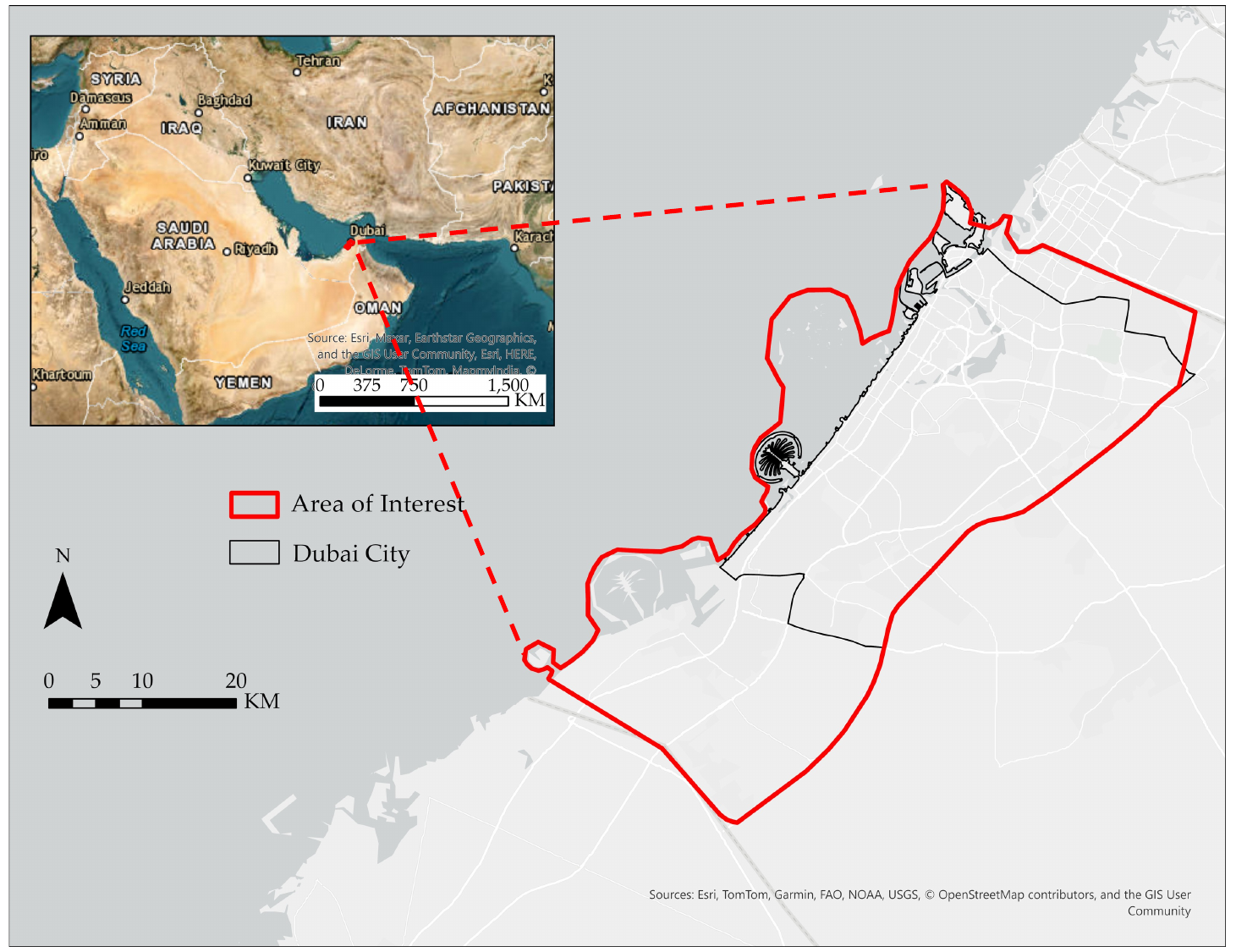}
    \caption{The extent of the real-world flood mapping study area in this paper, which covers around 1, 535 $\text{km}^2$ area along the southeastern coast of the Persian Gulf. The content in the red dot is shown in the highlighted area.}
    \label{fig:studyArea}
\end{figure}

\begin{figure*}[h!]
    \centering
    \includegraphics[width=0.99\linewidth]{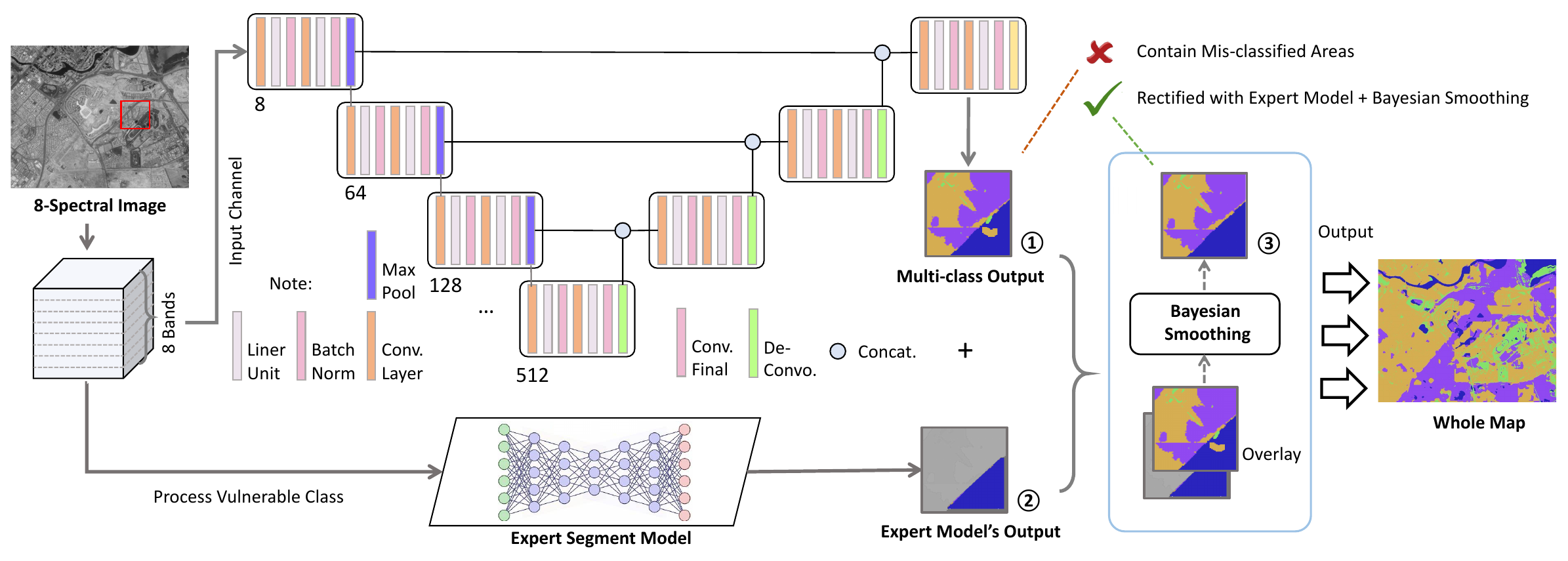}
    \caption{The framework of our method. It contains three components: Multi-class (m) U-net model that produces preliminary output in step \ding{172}, an Expert Segmentation model specifically for vulnerable class as output in \ding{172}, and a Bayesian Smoothing module for image semantic based boundary process to improve the different land-use cut by its context information. The expert segment model uses the only singular class of data as the training in Multi-class(m) U-net model, and yield better result by avoiding the confusions items in dataset, and the Bayesian Smoothing guarantees a context-based boundary among various classes in the flood mapping image. The image above shows the sampled processing in the red area, and the same procedure applies to the whole 8 spectral channel map. In the output \ding{172} contains mis-classified areas, which can be rectified by expert model in \ding{173}, and the \ding{174} provides better boundary differentiations.}
    \label{fig:mainfig}
\end{figure*}

To address these limitations, this paper presents \textbf{FM-LC}, a hierarchical framework for \textbf{\underline{F}}lood \textbf{\underline{M}}apping by \textbf{\underline{L}}and \textbf{\underline{C}}over identification. It builds upon a previously validated U-Net model \cite{hong2025forthcoming} with additional methodological advancements. It mainly comprises a three-stage architecture: we \textbf{First} produce a multi-class U-net model for segmentation of water, vegetation, built-area, and bare ground areas from multi-temporal PlanetScope mosaics. Inspection of these outputs often uncovers subtle misclassification, most frequently between water and moist vegetation, which we (\textbf{Second}) address by training a lightweight binary `expert' segmentation model dedicated to disentangling the troublesome class from all others. To polish the final maps, we then (\textbf{Third}) employ a Bayesian smoothing algorithm that draws on local neighborhood information to sharpen edges and suppress isolated noise.  We demonstrate the real-world effectiveness by applying it to the April 2024 Dubai flood event (in the area of Fig.~\ref{fig:studyArea}). This multi-layered strategy yields over 12\% average F1-score improvement and markedly crisper inundation boundaries versus conventional single-stage U-Net models.

\section{\textbf{Data and Preprocessing}}
The study area includes Dubai City and its surrounding metropolitan region, which covers approximately 1, 535 km\textsuperscript{2} along the southeastern coast of the Persian Gulf (Fig.~\ref{fig:studyArea}). It consists of six residential communities with a total estimated population of around 3.6 million in 2023~\cite{dsc2025population}. PlanetScope multispectral imagery employed for this study consists of eight spectral bands at a 3 m spatial resolution~\cite{planetlabs2024specs}. The rainfall event occurred from 15 to 17 April 2024; imagery was acquired on multiple dates before and after the storm, including 14 April (pre-rainfall), and composites for 18–19, 20–21, 25–26, and 27–28 April to ensure full coverage.

The labels for training data were derived from 2023 open‐access Sentinel-2 data at 10 m resolution~\cite{esa2018sentinel2} and manually corrected via the 14 April PlanetScope mosaic. Four classes were defined: water, vegetation, built area, and bare ground. The labeled dataset was cut into 1, 188 image patches (256 × 256 px, stride 128) for model training.

\section{\textbf{Hierarchical Deep Learning Framework}}

In this section, we describe the proposed hierarchical framework \textbf{FM-LC}, which operates in three sequential stages using a U-Net backbone for robust flood‐extent mapping.  Our pipeline begins by normalizing the raw eight-band PlanetScope input, followed by (1) a multi-class U-Net for coarse land-cover segmentation, (2) an expert binary classification model to refine the most confused classes, and (3) a Bayesian smoothing step to enforce spatial consistency and produce the final map.  Fig.~\ref{fig:mainfig} illustrates the overall workflow of our method, we begin with the Multi-class U-net backbone.

\subsection{The Backbond U-net Model}

Since U-Net has been widely used in image segmentation and classification tasks, we consider it as a powerful backbone. We denote the standardized input by 
\(\mathbf{X} = (\mathbf{X}_{\mathrm{raw}} - \boldsymbol{\mu}) \,/\, \boldsymbol{\sigma} \in \mathbb{R}^{H\times W\times 8}\),
where \(\mathbf{X}_{\mathrm{raw}}\) is the raw eight-band tile and \(\boldsymbol{\mu},\boldsymbol{\sigma}\in\mathbb{R}^8\) are the per-band means and standard deviations. Assume the multi-class U-Net is represented as a network $f$ with parameters $\theta$, we have $f_{\theta}$: 
\begin{equation}
    f_{\theta}:\mathbb{R}^{H\times W\times8}\longrightarrow[0,1]^{H\times W\times C}
\end{equation}
which indicates our network takes as input an $H\times W$ image with 8 spectral bands, i.e.\ a tensor in $\mathbb{R}^{H\times W\times8}$, and produces as output an $H\times W\times C$ tensor whose entries lie in $[0,1]$.  Concretely, for each pixel location $(i,j)$, the vector represents a probability distribution over the $C$ land-cover classes\footnote{Clarify: Here $i$ and $j$ are the pixel id, and `:' means apply to all channels.}:

\begin{equation}
    \bigl[f_{\theta}(\mathbf{X})\bigr]_{i,j,:}\in[0,1]^C
\end{equation}
We adapt a ResNet-34 encoder by replacing the initial convolution with the following configuration for our eight-band features: 
\begin{equation}
    \mathrm{Conv2d}(8,64,\,\mathrm{kernel}=7,\,\mathrm{stride}=2,\,\mathrm{padding}=3)
\end{equation}
and then attach a symmetric decoder with skip connections.  Training minimizes the multi-class cross-entropy as below:
\begin{equation}
    \mathcal{L}_{\mathrm{MC}}(\theta)= -\sum_{i=1}^{HW}\sum_{c=1}^{C}y_{i,c}\log p_{i,c} + \lambda\|\theta\|_2^2
\end{equation}

where \(c\) indexes the \(C\) classes.  The term \(-\,y_{i,c}\log p_{i,c}\) penalizes low predicted probability on the true class (since \(y_{i,c}=1\) only for the correct class), and the regularizer \(\lambda\|\theta\|_2^2\) (with \(\lambda>0\)) discourages excessively large weights in \(\theta\), improving generalization. Given the $p_{i,c}$, the coarse segmentation map is calculated as below:
\begin{equation}
\hat{Y}^{\mathrm{MC}}_i=\arg\max_c\,p_{i,c}
\end{equation}
by this, for each pixel $i$ we pick the single class $c$ whose predicted probability $p_{i,c}$ is largest, which means the model’s most confident choice, then we convert the soft probability distribution at each pixel into a hard label map where each pixel is assigned exactly one class. Even though we are able to train such a model with good convergence results, however, we still find it challenging to capture the nuanced difference between Water and Vegetation.

\subsection{\textbf{Expert Model for Specific Class Enhancement}}
For each hard class \(k\) (e.g., Water) that the coarse model confuses with partner \(k'\) (Vegetation), we train a binary U-Net which is represented by: 
\begin{equation}
  g_{\phi_k}:\mathbb{R}^{H\times W\times8}\longrightarrow[0,1]^{H\times W}  
\end{equation}
on the subset
\(\Omega_k=\{i:\hat{Y}^{\mathrm{MC}}_i\in\{k,k'\}\}\) by minimizing binary cross-entropy as follows: 
\begin{align}
\mathcal{L}_{\mathrm{EX},k}(\phi_k)
&= -\sum_{i\in\Omega_k}\bigl[y_{i,k}\log g_{\phi_k}(\mathbf{X})_i \\
&\quad+(1-y_{i,k})\log\bigl(1 - g_{\phi_k}(\mathbf{X})_i\bigr)\bigr].
\end{align}
where the \(y_{i,k}\in\{0,1\}\) indicates whether pixel \(i\) truly belongs to class \(k\) (\(y_{i,k}=1\)) or its confusing partner \(k'\) (\(y_{i,k}=0\)), and \(g_{\phi_k}(\mathbf{X})_i\in[0,1]\) is the expert U-Net’s predicted probability for class \(k\) at pixel \(i\).  When \(y_{i,k}=1\), the \(-\log g_{\phi_k}(\mathbf{X})_i\) penalizes model's low confidence on true positives; while \(y_{i,k}=0\), the term \(-\log\bigl(1 - g_{\phi_k}(\mathbf{X})_i\bigr)\) penalizes false positives.  This objective allows the expert to specialize on class \(k\).  At inference, thresholding at \(\tau_k\approx0.5\) yields a mask \(m_k(i)\) used to override the coarse labels:
\begin{equation}
\hat{Y}^{\mathrm{EX}}_i=
\begin{cases}
k,&m_k(i)=1,\\
k',&m_k(i)=0\ \land\ \hat{Y}^{\mathrm{MC}}_i\in\{k,k'\},\\
\hat{Y}^{\mathrm{MC}}_i,&\text{otherwise.}
\end{cases}    
\end{equation}

Despite the expert module’s ability in learning an accurate discrimination in \(\Omega_k\), the straightforward way that directly overwriting the coarse map can introduce abrupt boundary artifacts and isolated mislabels at the interface between multi-class and expert regions. To mitigate these `cut' issues, we apply a subsequent smoothing step that blends the two outputs and enforces spatial coherence, as described in next section.

% \subsection{\textbf{Bayesian Smoothing and Final Integration}}
% We enforce local spatial coherence on the expert-refined map by applying a Bayesian update to the logits
% \(\ell_{i,c}=\log\frac{p_{i,c}}{1-p_{i,c}}\).  Neighborhoods of size \(W\times W\) are extracted via \texttt{unfold}, and within each, the top \(k=\lceil W^2\alpha\rceil\) values yield mean \(m_{i,c}\) and variance \(s^2_{i,c}\).  The updated logit
% \[
% \ell'_{i,c}=\frac{s^2_{i,c}}{\sigma^2+s^2_{i,c}}\,m_{i,c}+\frac{\sigma^2}{\sigma^2+s^2_{i,c}}\,\ell_{i,c}
% \]
% and subsequent softmax produce smoothed probabilities \(p'_{i,c}\).  The final segmentation is
% \[
% \hat{Y}^{\mathrm{SM}}_i=\arg\max_c\,p'_{i,c}.
% \]

\subsection{\textbf{Bayesian Smoothing and Final Integration}}
According to the previous study~\cite{weller2011bayesian}, it is proved that using pixel neighborhood information and a prior assumption on class distributions, we could perform a posterior probabilistic smoothing of each pixel’s label. A common approach is to treat the entire segmentation map as a Markov Random Field (MRF), assuming neighboring pixels are more likely to share the same label, and then use Bayesian (or maximum a posteriori) inference to re-estimate each pixel’s class.

In a full MRF formulation, one would minimize the global energy:
\begin{equation}
    E(L) \;=\; -\sum_{i}\log\pi_i(L_i)\;+\;\beta\sum_{(i,j)\in\mathcal{N}}w_{ij}\,\mathbf{1}[L_i\neq L_j]
\end{equation}
but solving this exactly at high resolution is computationally expensive. Instead, we propose to approximate the likelihood-prior tradeoff with a localized Bayesian update on the logit tensor
\(\ell_{i,c}=\log\frac{p_{i,c}}{1-p_{i,c}}\). For each class channel, we first pad the logit map to handle border pixels, then slide a \(W\times W\) window across the padded map to gather all overlapping neighborhoods of logits at each location, and select the top \(k = \lceil W^2\,\alpha\rceil\) logit values from each neighborhood. After this process, the empirical mean \(m_{i,c}\) could serve as a local `prior' and the variance \(s^2_{i,c}\) could measure neighbor agreement. Interpreting \(\ell_{i,c}\) as a Gaussian observation with noise variance \(\sigma^2\) and \(m_{i,c}\) as a Gaussian prior, then we get the posterior mode value is as below: 
\begin{equation}
\ell'_{i,c}
=\frac{s^2_{i,c}}{\sigma^2 + s^2_{i,c}}\,m_{i,c}
+\frac{\sigma^2}{\sigma^2 + s^2_{i,c}}\,\ell_{i,c},
\end{equation}
which automatically down‐weights the prior when neighborhood variance is high and enforces consistency when neighbors strongly agree, this provides contextual dependency and better splits the classes based on the perceived image semantics. Applying softmax to \(\ell'\) yields smoothed probabilities \(p'_{i,c}\) that blend each pixel’s own confidence with its local context. The final segmentation map can be represented as below: 
\begin{equation}
\hat{Y}^{\mathrm{SM}}_i = \arg\max_c\,p'_{i,c},    
\end{equation}
and the generated map is then saved as a \texttt{uint8} \texttt{GeoTIFF} for downstream analysis. By three complementary steps, the \textbf{FM-LC} can eventually achieve high accuracy on high-resolution and fine-grained features. For a better understanding, we provide the detailed workflow in the pseudo code as below: 

% \begin{algorithm}[!ht]
% \caption{Hierarchical U-Net Segmentation}\label{alg:hier}
% \begin{algorithmic}[1]
% \Require 8-band image \(\mathbf{X}\in\mathbb{R}^{H\times W\times8}\)
% \Ensure Final segmentation \(\hat{Y}^{\mathrm{SM}}\)
% \State Compute \(P\gets f_{\theta}(\mathbf{X})\) and \(\hat{Y}^{\mathrm{MC}}_i\gets\arg\max_c P_{i,c}\)
% \State Identify confusing classes set \(\mathcal{S}=\{k:\text{confused with }k'\}\)
% \State Initialize \(\hat{Y}^{\mathrm{EX}}\gets\hat{Y}^{\mathrm{MC}}\)
% \ForAll{\(k\in\mathcal{S}\)}
%   \State Compute expert mask \(M_k\gets g_{\phi_k}(\mathbf{X})\)
%   \ForAll{pixels \(i\) with \(\hat{Y}^{\mathrm{MC}}_i\in\{k,k'\}\)}
%     \If{\(M_k(i)\ge\tau_k\)}
%       \State \(\hat{Y}^{\mathrm{EX}}_i\gets k\)
%     \Else
%       \State \(\hat{Y}^{\mathrm{EX}}_i\gets k'\)
%     \EndIf
%   \EndFor
% \EndFor
% \State Extract neighborhoods, compute \(m_{i,c},s^2_{i,c}\) via unfold and top-\(k\)
% \State Update logits \(\ell'_{i,c}=\tfrac{s^2_{i,c}}{\sigma^2+s^2_{i,c}}\,m_{i,c}+\tfrac{\sigma^2}{\sigma^2+s^2_{i,c}}\,\ell_{i,c}\)
% \State Compute smoothed probabilities \(p'_{i,c}=\mathrm{softmax}(\ell')\)
% \State \(\hat{Y}^{\mathrm{SM}}_i\gets\arg\max_c p'_{i,c}\)
% \State Save \(\hat{Y}^{\mathrm{SM}}\) as uint8 GeoTIFF
% \Return \(\hat{Y}^{\mathrm{SM}}\)
% \end{algorithmic}
% \end{algorithm}
\begin{algorithm}[!ht]
\caption{Hierarchical \textbf{FM-LC} Segmentation}\label{alg:hier}
\begin{algorithmic}[1]
\Require 8-band image \(\mathbf{X}\)
\Ensure Final segmentation \(\hat{Y}^{\mathrm{SM}}\)
\State Compute \(P=f_{\theta}(\mathbf{X})\) and \(\hat{Y}^{\mathrm{MC}}=\arg\max_c P\)
\State Detect confusing classes \(\mathcal{S}\) and initialize \(\hat{Y}^{\mathrm{EX}}\gets\hat{Y}^{\mathrm{MC}}\)

\ForAll{\(k\in\mathcal{S}\)}
  \State \(M_k\gets g_{\phi_k}(\mathbf{X})\); for each \(i\) with \(\hat Y^{\mathrm{MC}}_i\in\{k,k'\}\), set
  \(\hat Y^{\mathrm{EX}}_i\gets k\) if \(M_k(i)\ge\tau_k\) else \(k'\)
\EndFor
\State Extract \(W\times W\) neighborhoods via \texttt{unfold}, compute local mean \(m\) and variance \(s^2\)
\State Update logits \(\ell'\) by blending with \(m,s^2\), compute \(p'=\mathrm{softmax}(\ell')\)
\State \(\hat{Y}^{\mathrm{SM}}=\arg\max_c p'\), save as GeoTIFF
\end{algorithmic}
\end{algorithm}

\begin{table*}[!t]
  \caption{Evaluation comparison between the baseline U-Net model, the multi-class U-Net with Bayesian optimization, and the improved Hierarchical U-Net. The ($\uparrow$) means the metrics value, the higher the better.}
  \centering
  \resizebox{0.90\linewidth}{!}{%
  \begin{tabular}{@{}l*{3}{ccc}@{}}
    \toprule
    \multirow{2}{*}{Land Class}
      & \multicolumn{3}{c}{DICE ($\uparrow$)}
      & \multicolumn{3}{c}{Precision ($\uparrow$)}
      & \multicolumn{3}{c}{F1-Score ($\uparrow$)} \\
    \cmidrule(lr){2-4} \cmidrule(lr){5-7} \cmidrule(lr){8-10}
      & UNet(m) & UNet+Bayes & Hierarchical
      & UNet(m) & UNet+Bayes & Hierarchical
      & UNet(m) & UNet+Bayes & Hierarchical \\
    \midrule
    Water         & 0.918 & 0.921 & \(\mathbf{0.921}_{\scriptstyle+0.002}\)
                  & 0.953 & 0.976 & \(\mathbf{0.976}_{\scriptstyle+0.012}\)
                  & 0.919 & 0.924 & \(\mathbf{0.924}_{\scriptstyle+0.003}\) \\
    Vegetation    & 0.641 & 0.558 & \(\mathbf{0.896}_{\scriptstyle+0.297}\)
                  & 0.865 & 0.919 & \(\mathbf{0.929}_{\scriptstyle+0.037}\)
                  & 0.698 & 0.658 & \(\mathbf{0.913}_{\scriptstyle+0.235}\) \\
    Built area    & 0.874 & 0.898 & \(\mathbf{0.905}_{\scriptstyle+0.019}\)
                  & 0.906 & 0.933 & \(\mathbf{0.950}_{\scriptstyle+0.031}\)
                  & 0.905 & 0.902 & \(\mathbf{0.910}_{\scriptstyle+0.007}\) \\
    Bare ground   & 0.765 & 0.771 & \(\mathbf{0.772}_{\scriptstyle+0.004}\)
                  & 0.813 & 0.853 & \(\mathbf{0.858}_{\scriptstyle+0.025}\)
                  & 0.855 & 0.867 & \(\mathbf{0.870}_{\scriptstyle+0.009}\) \\
    \bottomrule
  \end{tabular}%
  }\label{tab:threemetrics}
\end{table*}

\begin{figure*}
    \centering
    \includegraphics[width=0.99\linewidth]{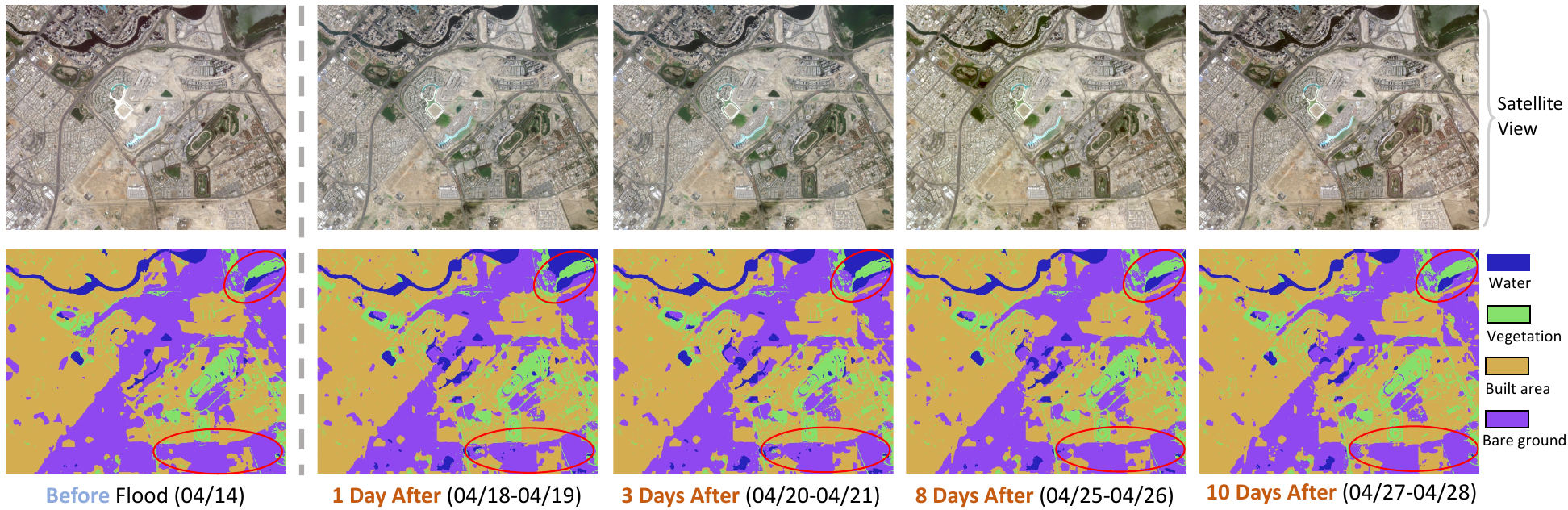}
    \caption{The qualitative results with visual presentations: We illustrate the detected land cover status by two rows: satellite view, and our model's segmentation result. It is obvious that the output by our framework provides a more straightforward way to track the flood (water) area changes. By comparing before the flood event and various days after the event, we can observe that the area of the flood is getting larger comparing the 1 Day after the event to the date 04/14,  and is getting smaller as time passes. The red highlighted areas are the local places easier to observe the change.}
    \label{fig:Fig3}
\end{figure*}

\section{Experimental Results}
In this section, we conducted experiments to verify the proposed method's effectiveness. We divide the experiments into two types: \textit{Quantitative Evaluation} and \textit{Qualitative Analysis}.
\begin{figure}[h!]
    \centering
    \includegraphics[width=0.7\linewidth]{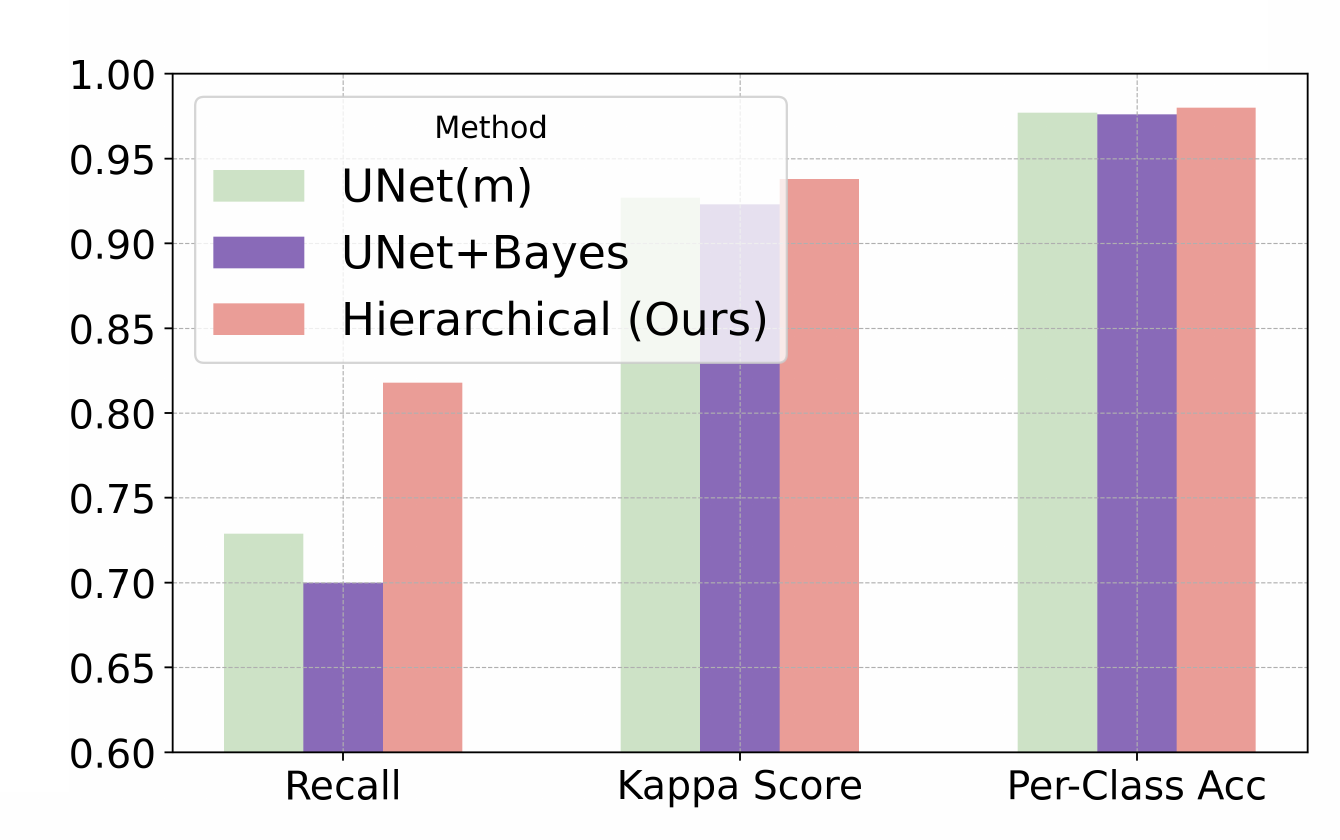}
    \caption{The comparison on metrics: Recall, Kappa Score, and Per-class Accuracy, which shows the high performance of \textbf{FM-LC} Hierarchical framework.}
    \label{fig:bars}
    \vspace{-3mm}
\end{figure}
\vspace{-3mm}
\subsection{Quantitative Evaluation} 
In our experiments, we assembled a real‐world flood dataset covering ~1, 535 $\text{km}^2$ urban areas in the Dubai metropolitan region. All imagery and labels were tiled into $256\times256$\,px patches, then randomly split into training (70\%) and testing (30\%) subsets using a fixed seed (42) to ensure reproducibility. Then the Expert (binary) model was trained to perform the first round multi-class segmentation, then the vulnerable class can be identified by the sampled test-set, based on this, the class‐specific `expert' can be trained leveraging the same training data but specifically for the identified class. Bayesian smoothing is then applied. To produce valid evaluations, we adopted the \textbf{metrics} following existing work~\cite{wang2020image}, including: recall, precision, F1‐score, confusion matrices, overall Kappa~\cite{brennan1981coefficient}, and accuracy, which were all conducted on the held‐out 30\% test set. As shown in Table.~\ref{tab:threemetrics}, on three widely used metrics, our method (Hierarchical) framework consistently shows better performance than other baselines such as Pure U-Net models with \underline{m}ulti-class heads - UNet(m) and UNet(m) with Bayesian smoothing. The UNet(m) alone is worse than our method, mainly in the \texttt{vegetation} class, this is because the multi-class U-net method can hardly learn well in distinguishing the similar classes (Water and Vegetation), and our method helps by an expert small model for vegetation using the same amount of the data. Another three metrics, such as Recall, Kappa score, and Per-class Accuracy, are also visible in Fig.~\ref{fig:bars}, it is also clearly represented that the Recall and Kappa Score shows strong improvement, while the improvement on per‐class ACC is not obvious, this is because a poorly performing class \(k\) occupies only a small fraction of the total pixels.  Specifically, the overall accuracy can be written as
\(\mathrm{ACC}_{\mathrm{overall}} = \frac{1}{HW}\sum_{i=1}^{H}\sum_{j=1}^{W}\mathbf{1}\{\hat{Y}_{i,j}=Y_{i,j}\} 
= \sum_{c=1}^{C}\underbrace{\frac{\bigl|\{(i,j)\colon Y_{i,j}=c\}\bigr|}{HW}}_{w_c}\,\mathrm{ACC}_c,\)
where each weight \(w_c\) is the proportion of pixels belonging to class \(c\). If class \(k\) has a small \(w_k\), then even a low \(\mathrm{ACC}_k\) contributes little to \(\mathrm{ACC}_{\mathrm{overall}}\), causing the overall accuracy to be dominated by well-performing classes that occupy most of the area.

Lastly, we also showcase the confusion matrix of the hierarchical \textbf{FM-LC} model. The high overall accuracy and Kappa statistics make it a practical approach for real-world flood trajectory analysis. In conclusion, despite the overall accuracy showing non-significant improvement, all other evaluation metrics prove that our method, Hierarchical \textbf{FM-LC}, makes consistent advancement in the urban flood
mapping task.

\begin{table*}[!t]
  \centering
  \caption{The Confusion matrix results of the Hierarchical \textbf{FM-LC} model}
  \label{tab:confusion_matrix}
  \resizebox{0.70\linewidth}{!}{%
  \begin{tabular}{@{}lrrrrrrr@{}}
    \toprule
    & \multicolumn{4}{c}{Reference data} 
    & \multicolumn{1}{c}{Row Total} 
    & \multicolumn{1}{c}{Producer accuracy (\%)} \\
    \cmidrule(lr){2-5} \cmidrule(lr){6-6} \cmidrule(lr){7-7}
    Classified data 
      & Water      & Vegetation  & Built Area   & Bare Ground 
      & Total      & \% \\ 
    \midrule
    Water       & \cellcolor{gray!20}2,532,004 
                & 2,158   
                & 28,044   
                & 20,066   
                & 2,582,272 
                & 98.33 \\
    Vegetation  & 121  
                & \cellcolor{gray!20}691,170   
                & 11,840   
                & 4,077   
                & 707,208 
                & 97.73 \\
    Built Area  & 33,406  
                & 34,986   
                & \cellcolor{gray!20}7,705,119  
                & 262,471   
                & 8,035,982 
                & 95.88 \\
    Bare Ground & 7,840  
                & 16,966   
                & 187,956   
                & \cellcolor{gray!20}4,059,344  
                & 4,272,106 
                & 95.02 \\
    \midrule
    Column Total& 2,573,371 
                & 745,280   
                & 7,932,959  
                & 4,345,958  
                & 15,597,648 
                &       \\
    User accuracy (\%) 
                & 98.33  
                & 92.74   
                & 97.13   
                & 93.48   
                &           &       \\
    \midrule
    \textbf{Overall accuracy (\%)} 
                & \multicolumn{6}{r}{96.09} \\
    \textbf{Kappa statistic}       
                & \multicolumn{6}{r}{0.939} \\
    \bottomrule
  \end{tabular}
  }
  \vspace{-3mm}
\end{table*}

\subsection{Qualitative Analysis with Visual Presentation}
A side-by-side comparison between the raw PlanetScope imagery and corresponding classification results for a zoomed-in section is presented in Fig.~\ref{fig:Fig3}; it contains the classification results for 14 April (pre-rainfall) and the other 4 days after the rainfall. These highlight the model's ability to delineate class boundaries and detect subtle differences in the landscapes. We can see that even narrow water bodies and scattered vegetation patches are clearly identified. For a more specific case analysis, we could track the circled areas: the image before the Flood shows little water area in red circles, while after one day of significant rainfall, the water coverage increases most, after that three days, not obvious changes compared to one days after, this is because during the period, light rains still happens. Around one week later (after 8-10 days), the water area gradually falls back to the level before rainfall. The mapped results demonstrate the effectiveness of the proposed model in capturing flood extent across diverse urban and natural landscapes.

\vspace{-3mm}
\section{Conclusion}
In this paper, we have presented \textbf{FM-LC}, a three‐stage hierarchical framework that combines a multi‐class U‐Net backbone, class‐specific expert binary U‐Nets, and efficient Bayesian smoothing to resolve spectral ambiguities in arid urban flood mapping.  Applied to the April 2024 Dubai event, FM‐LC delivered over 12\% average F1 improvement and markedly sharper inundation outlines, while remaining fast enough for daily PlanetScope monitoring.  Its modular design naturally supports extensions, such as multi‐temporal analysis, coupling with hydrodynamic models, or domain‐adaptation strategies, to tackle diverse flood‐mapping challenges.  We hope FM-LC will serve as a versatile foundation and inspire the community to develop hybrid deep learning pipelines for resilient, data-driven environmental mapping across diverse climatic and urban contexts similar to the Persian Gulf region.

% \section*{Acknowledgment}
% \textcolor{red}{The authors thank XYZ for support.}

\bibliographystyle{IEEEtran}
\bibliography{bibtex/bib/IEEEexample}
% \begin{thebibliography}{1}
% \bibitem{IEEEhowto:kopka}
% H.~Kopka and P.~W. Daly, \emph{A Guide to \LaTeX}, 3rd~ed., Addison-Wesley, 1999.
% \end{thebibliography}

% \begin{IEEEbiography}{Xin Hong}
% Biography text here.
% \end{IEEEbiography}

% \begin{IEEEbiography}{Longchao Da}
% Biography text here.
% \end{IEEEbiography}

\end{document}